\documentclass[letterpaper, 10 pt, conference]{ieeeconf}  

\IEEEoverridecommandlockouts                              

\usepackage{amsmath}
\usepackage{multirow}
\usepackage{makecell}

\usepackage{graphicx}
\usepackage{amssymb}
\usepackage[caption=false,font=footnotesize]{subfig}

\usepackage{cite}
\usepackage[table]{xcolor}   
\definecolor{hanpurple}{rgb}{0.32, 0.09, 0.98}
\makeatletter
\let\NAT@parse\undefined
\makeatother
\usepackage[pagebackref=false,breaklinks=true,colorlinks,bookmarks=false]{hyperref}
\hypersetup{colorlinks=true,
    linkcolor={red},
    citecolor={hanpurple},
    urlcolor={magenta}
}

\title{\LARGE \bf
Real-Time LiDAR Super-Resolution via Frequency-Aware Multi-Scale Fusion
}

\author{June Moh Goo$^{*}$, Zichao Zeng and Jan Boehm
\thanks{The authors are with Department of Civil, Environmental and Geomatic Engineering,
        University College London, WC1E 6BT London, U.K.
        }
\thanks{This work was supported by the Engineering and Physical Sciences Research Council through an industrial CASE studentship with Ordnance Survey (Grant number EP/X524840/1 and EP/W522077/1).}
\thanks{*\; Corresponding Author: June Moh Goo. Email: june.goo.21@ucl.ac.uk}
}

\begin{document}

\maketitle
\thispagestyle{empty}
\pagestyle{empty}

\begin{abstract}
 LiDAR super-resolution addresses the challenge of achieving high-quality 3D perception from cost-effective, low-resolution sensors. While recent transformer-based approaches like TULIP show promise, they remain limited to spatial-domain processing with restricted receptive fields. We introduce FLASH (Frequency-aware LiDAR Adaptive Super-resolution with Hierarchical fusion), a novel framework that overcomes these limitations through dual-domain processing. FLASH integrates two key innovations: (i) Frequency-Aware Window Attention that combines local spatial attention with global frequency-domain analysis via FFT, capturing both fine-grained geometry and periodic scanning patterns at log-linear complexity. (ii) Adaptive Multi-Scale Fusion that replaces conventional skip connections with learned position-specific feature aggregation, enhanced by CBAM attention for dynamic feature selection. Extensive experiments on KITTI demonstrate that FLASH achieves state-of-the-art performance across all evaluation metrics, surpassing even uncertainty-enhanced baselines that require multiple forward passes. Notably, FLASH outperforms TULIP with Monte Carlo Dropout while maintaining single-pass efficiency, which enables real-time deployment. The consistent superiority across all distance ranges validates that our dual-domain approach effectively handles uncertainty through architectural design rather than computationally expensive stochastic inference, making it practical for autonomous systems.
\end{abstract}

\section{Introduction}
The high cost of high-resolution LiDAR sensors presents a fundamental challenge for autonomous systems. LiDAR (Light Detection and Ranging) sensors have become essential for autonomous driving \cite{auto1, auto2, auto3}, robotics \cite{robo1, robo2}, and 3D scene understanding \cite{3Dscene1} due to their ability to capture precise spatial information. While these sensors enable critical tasks such as object detection and semantic segmentation, the demand for increasingly detailed perception continues to push the boundaries of sensor resolution.

However, high-resolution LiDAR sensors face practical deployment barriers. They are prohibitively expensive, require substantial memory and computational resources, and increase bandwidth requirements that challenge real-time processing capabilities \cite{chen2025srmamba, yang2024tulip}. Moreover, unlike traditional computer vision for images, where models can often transfer across different camera resolutions through simple pre-processing, LiDAR presents unique challenges for cross-resolution adaptation: the discrete beam patterns, non-uniform point distributions, and sensor-specific scanning geometries require training separate models for each sensor configuration, which is highly impractical \cite{review1}. These limitations restrict the deployment of high-resolution LiDAR in cost-sensitive or resource-constrained environments.

\begin{figure}[t]
    \centering
    \includegraphics[width=\linewidth]{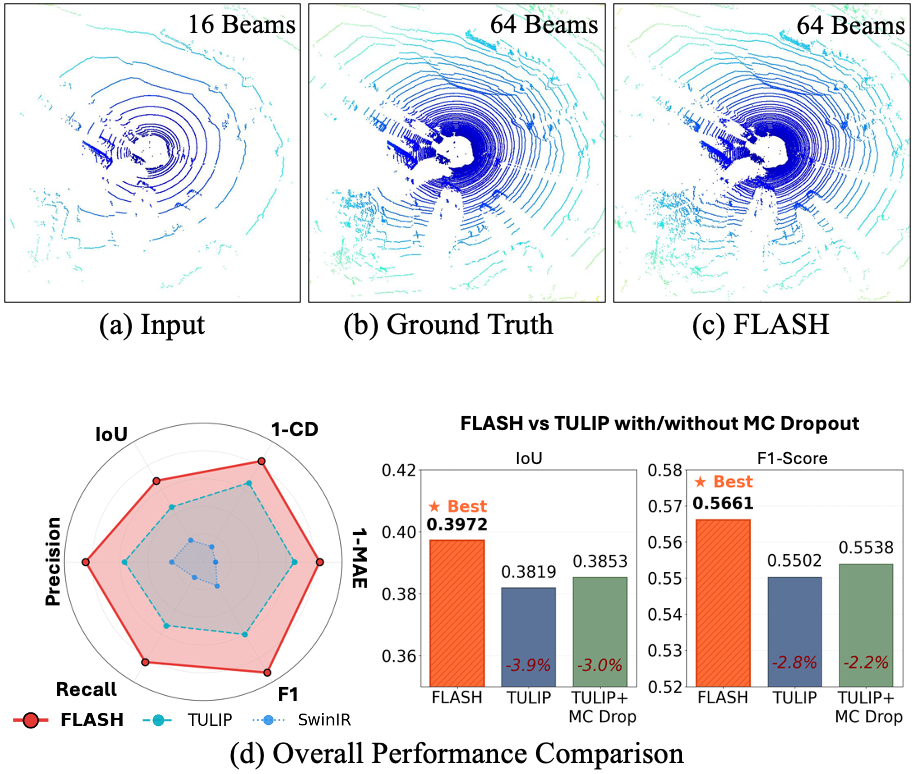}
    \caption{Overview of FLASH performance. (a–c) Visual comparison of range image super-resolution (16×1024 → 64×1024). (d) Left: radar chart comparing performance metrics against competing methods on KITTI dataset (each metric with custom scale normalization). Right: bar graphs demonstrate that even with MC Dropout enhancement, TULIP variants fail to achieve FLASH's IoU and F1 score performance.}
    \label{fig:intro}
\end{figure}


LiDAR super-resolution techniques have emerged as a promising solution to bridge this gap \cite{yang2024tulip, lidar_sr, iln}. By enhancing the spatial resolution of range images from affordable low-resolution sensors, these methods aim to achieve performance comparable to expensive high-resolution devices while maintaining manageable costs and resource requirements. The diversity of LiDAR sensors in real-world deployments, characterized by varying resolutions, fields of view, and scanning patterns, presents a challenge. This diversity requires robust super-resolution models that can generalize across sensor configurations without retraining.

Recent advances have made significant progress towards these goals. LiDAR-SR converted the 3D upsampling problem into 2D image super-resolution using simulation-based training and Monte Carlo dropout for uncertainty estimation, enabling cross-sensor generalization \cite{lidar_sr}. TULIP \cite{yang2024tulip} achieved state-of-the-art results through a Swin Transformer \cite{liu2021swin, liang2021swinir}-based U-Net architecture with window attention mechanisms, demonstrating robust performance across different datasets. However, these methods face fundamental limitations: they operate with restricted receptive fields, process features exclusively in the spatial domain, and rely on computationally expensive uncertainty quantification (e.g., Monte Carlo Dropout with 10× computational overhead - 20 forward passes processed in batches) to achieve robust predictions.

\begin{figure*}[t]
    \centering
    \includegraphics[width=\textwidth]{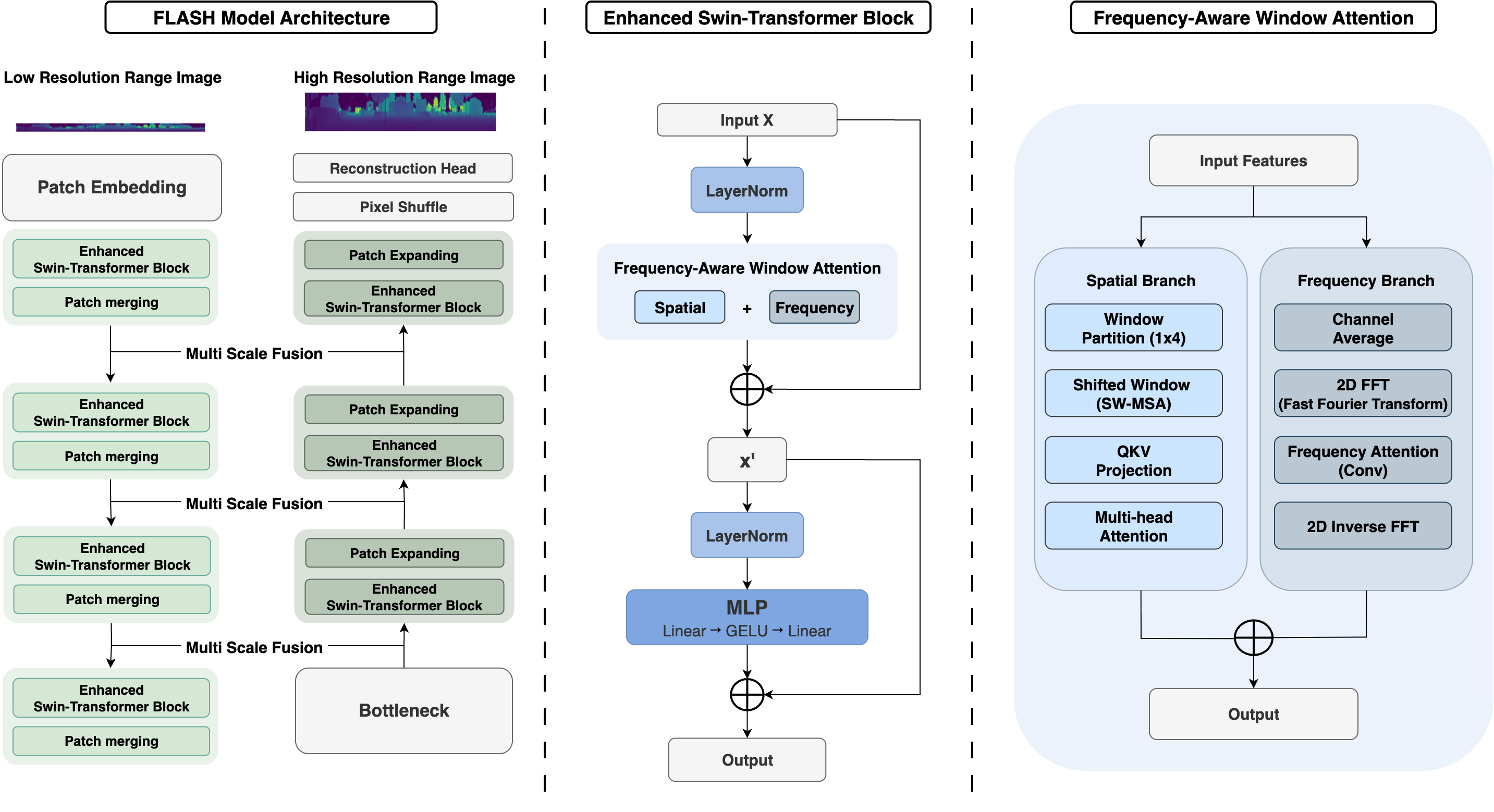}
    \caption{FLASH architecture overview. The encoder-decoder network processes low-resolution range images (16×1024) to produce high-resolution outputs (64×1024) through Enhanced Swin-Transformer blocks with Frequency-Aware Window Attention (FA) and Multi-Scale Fusion (MSF) at skip connections. The FA module (right) employs dual-branch processing combining spatial window attention with frequency domain analysis via FFT.}
    \label{fig:FLASH_FA}
\end{figure*}

Figure~\ref{fig:intro} provides an overview of FLASH's performance, demonstrating superior results compared to existing methods, even those enhanced with uncertainty quantification. In this paper, we introduce FLASH, a novel LiDAR super-resolution framework that overcomes current limitations through a fundamentally different approach. Rather than relying on computationally expensive uncertainty quantification methods, FLASH achieves robust reconstruction through two synergistic architectural innovations. The Frequency-Aware Window Attention mechanism operates in dual domains simultaneously by capturing fine-grained local geometry through spatial attention while extracting global scanning patterns via frequency analysis. This dual-domain processing expands the effective receptive field beyond window boundaries without additional computational burden. Meanwhile, our Adaptive Multi-Scale Fusion module learns to weight features dynamically based on spatial position and scale, moving beyond the fixed concatenation used in existing methods. Together, these components enable FLASH to surpass uncertainty-enhanced baselines while maintaining real-time efficiency. Our main contributions are:

\begin{itemize}
\item \textbf{A novel dual-domain attention mechanism} that processes both spatial and frequency information significantly expands the effective receptive field while preserving local detail through parallel branch processing.

\item \textbf{An adaptive multi-scale fusion module} that replaces conventional skip connections with learned position-specific aggregation of multi-resolution features, enhancing the preservation of geometric detail in the decoder path.

\item \textbf{State-of-the-art results on KITTI} across all metrics, including MAE, Chamfer Distance, IoU, and F1-score, validated through extensive experiments.

\item \textbf{Robust performance across all distance ranges}, maintaining accuracy from near to far regions where existing methods typically degrade, demonstrating the effectiveness of our architectural approach.

\item \textbf{Superior performance without uncertainty quantification}, outperforming TULIP with Monte Carlo Dropout while requiring only single-pass inference, enabling real-time deployment.
\end{itemize}

\section{Related Work}

\subsection{Image Resolution using Deep Learning}
Recent transformer-based approaches have achieved remarkable results in image super-resolution. SwinIR \cite{liang2021swinir} introduced local window attention with shifted windows for efficient high-resolution image reconstruction, while HAT \cite{hat} enhanced this with channel attention and overlapping cross-attention to better activate pixels. These methods demonstrate the effectiveness of attention mechanisms for capturing long-range dependencies while maintaining computational efficiency.

However, these methods target RGB images with smooth textures and gradients, unlike LiDAR range images which contain sharp discontinuities, extreme aspect ratios (1:32 to 1:64), and single-channel depth where errors directly impact 3D geometry.

\subsection{Range Image-Based LiDAR Super-Resolution}
Several methods have specifically addressed LiDAR super-resolution through range image processing. LiDAR-SR pioneered this approach with a CNN-based U-Net architecture, introducing Monte Carlo dropout during inference to identify and remove uncertain predictions at object boundaries \cite{lidar_sr}. This uncertainty-aware approach effectively reduces noise but relies on convolutional operations that inherently smooth sharp discontinuities.

ILN took a different approach by learning interpolation weights rather than directly predicting range values, using an implicit neural representation to maintain geometric accuracy~\cite{iln}. While this method preserves input geometry well, it operates with limited neighbouring context, potentially missing larger structural patterns in the scene. Additionally, the method was tested only on synthetic training data, raising questions about its robustness to real-world sensor noise and irregular scanning patterns.

TULIP \cite{yang2024tulip} advanced the field by adapting Swin Transformer for range images, introducing row-based patching (1×4) to preserve vertical resolution and non-square windows for multi-scale attention. The circular padding in the horizontal dimension naturally accommodates the 360 degree field of view of rotating LiDARs. These design choices demonstrate the importance of architecture modifications specific to range image characteristics. While TULIP can employ Monte Carlo Dropout for uncertainty estimation, this requires multiple forward passes that significantly increase computational cost.

Despite these advances, existing methods operate solely in the spatial domain, missing global patterns inherent in LiDAR scanning. When robustness is needed, they rely on computationally expensive stochastic inference (Monte Carlo Dropout). These limitations motivate exploring frequency domain processing and adaptive fusion strategies for inherently robust predictions.

\subsection{Frequency Domain Processing}
Frequency domain analysis through Fast Fourier Transform (FFT) has emerged as an efficient mechanism for capturing global context in vision tasks \cite{ffc, ffc_inpaint, sinha2022nl_ffc}. FFC (Fast Fourier Convolution) directly replaces spatial convolutions with FFT operations, achieving full-image receptive fields with O(n log n) complexity for dense prediction tasks \cite{ffc}. AFNO (Adaptive Fourier Neural Operators) \cite{afno} demonstrates that FFT-based token mixing can replace self-attention in transformers while reducing computational costs from quadratic to log-linear.

For image processing, several architectures leverage frequency domain insights. GFNet (Global Filter Networks)~\cite{GFNet} employs FFT to learn global filters in frequency space, providing an efficient alternative to self-attention for long-range dependency modeling. FcaNet \cite{qin2021fcanet} introduces frequency channel attention, showing that different frequency components contain complementary information for visual recognition.

Recently, applying FFC to image super-resolution has gained attention. NL-FFC (Non-Local Fast Fourier Convolution) \cite{sinha2022nl_ffc} combines non-local attention with FFC specifically for image super-resolution, achieving competitive performance on standard benchmarks. Their work demonstrates that frequency features provide faster convergence on low-frequency components, which then serve as priors for unobserved high-frequency details. This validates the potential of frequency domain processing for reconstruction tasks.

LiDAR range images contain periodic scanning patterns and sharp depth discontinuities that are separated in frequency space. Low frequencies capture smooth surfaces, while high frequencies preserve edges. This motivates our dual-domain approach: spatial attention for local details and frequency analysis for global patterns, expanding receptive fields without losing precision.

\subsection{Multi-Scale Feature Fusion}
Multi-scale feature fusion has evolved from simple skip connections to sophisticated adaptive mechanisms. U-Net~\cite{unet} pioneered direct concatenation between encoder and decoder features, preserving fine details during upsampling. FPN \cite{FPN} advanced this concept with top-down pathway and lateral connections, while PANet \cite{PANet} added bottom-up path augmentation to enhance information flow. However, these methods apply uniform fusion across all spatial locations, regardless of content.

Recent works have explored dynamic fusion strategies. BiFPN \cite{tan2020efficientdet} introduced learnable weights for different input features, allowing the network to learn which features to emphasize at each scale. ASFF (Adaptively Spatial Feature Fusion) \cite{asff} proposed adaptive fusion at different spatial locations, recognizing that optimal feature combinations vary across the image.

Attention mechanisms have proven effective for feature refinement. CBAM \cite{woo2018cbam} sequentially applies channel and spatial attention modules, helping networks focus on informative features. This lightweight module has shown consistent improvements when integrated into various architectures, demonstrating that selective feature emphasis significantly enhances fusion quality.


Adaptive multi-scale fusion is essential for LiDAR range images due to diverse geometric patterns. Sharp boundaries need fine-scale features, while smooth surfaces benefit from broader context. We learn position-specific fusion weights across multiple kernel sizes (1x1, 3x3, 5x5) and incorporate CBAM to enable dynamic feature selection based on local content, thereby avoiding uniform fusion.

Overall, existing super-resolution methods face a fundamental trade-off. Robust inference often relies on stochastic sampling with high computational cost, while spatial-only processing overlooks global patterns in scanning data. To address this gap, we present FLASH, which combines dual-domain attention with adaptive fusion to deliver robust predictions without an additional computational burden.
\section{Methodology}
\begin{figure*}[t]
    \centering
    \includegraphics[width=\linewidth]{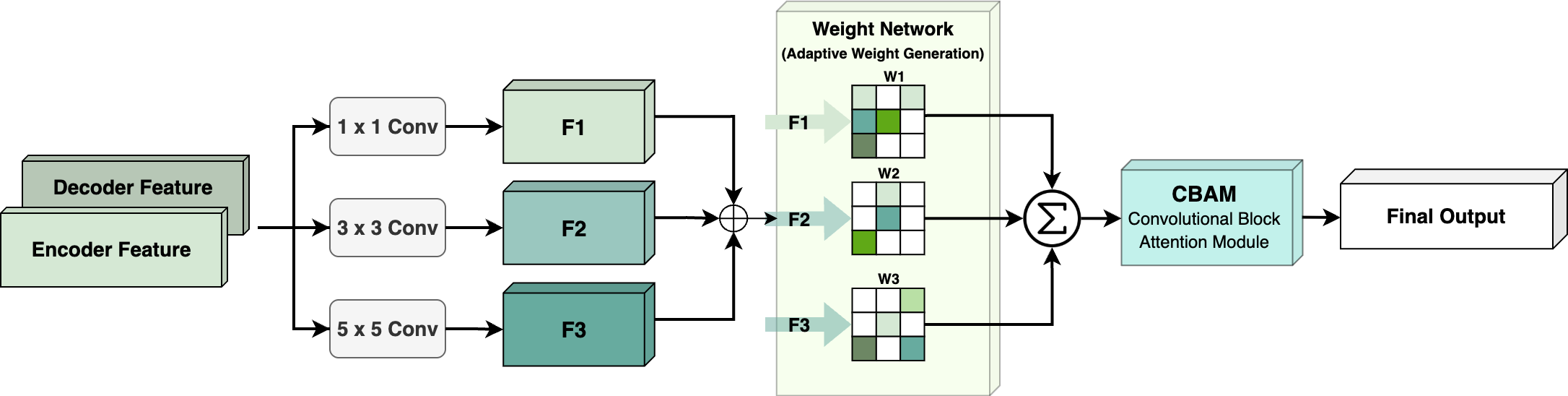}
    \caption{Multi-Scale Fusion (MSF) module. Encoder and decoder features are processed through parallel multi-scale convolutions (1×1, 3×3, 5×5). Adaptive weights are generated for position-specific fusion, followed by CBAM refinement for enhanced feature selection.}
    \label{fig:MSF}
\end{figure*}

\subsection{Problem Formulation}
Given a low-resolution LiDAR point cloud 
\(P_l \in \mathbf{R}^{n_l \times 3}\) 
captured by a sensor with \(H_l\) vertical channels, our goal is to generate a high-resolution point cloud 
\(P_h \in \mathbf{R}^{n_h \times 3}\) 
that approximates the output of a sensor with 
\(H_h = 4 \cdot H_l\) channels. Following established practice in range image-based LiDAR processing, we transform this 3D upsampling problem into a 2D image super-resolution task.  

We first project the point cloud onto a range image 
\(I_l \in \mathbf{R}^{H_l \times W}\) 
using spherical projection, where each pixel encodes the Euclidean distance 
\(r = \sqrt{x^2 + y^2 + z^2}\). The horizontal resolution \(W\) remains constant (\(W = 1024\) in our experiments) as we focus on vertical upsampling, consistent with the asymmetric resolution characteristics of rotating LiDARs. The projection follows:
\begin{equation}
    u = \frac{W}{2} - \frac{W}{2\pi} \arctan2\!\left(y, x\right),
\end{equation}
\begin{equation}
v = \frac{H}{\theta_{\max} - \theta_{\min}} \left(\theta_{\max} - \arctan\left(\frac{z}{\sqrt{x^2 + y^2}}\right)\right)
\end{equation}

where \(\theta_{\max}\) and \(\theta_{\min}\) define the vertical field of view. Finally, we apply a logarithmic transformation \(\log(r+1)\) to compress the range distribution, improving network convergence for distant points.

Our network predicts the high-resolution range image 
\(I_h \in R^{H_h \times W}\) 
from the low-resolution input 
\(I_l \in R^{H_l \times W}\), 
where \(H_h = 4H_l\) for \(4\times\) upsampling. We adopt TULIP's U-Net architecture with Swin Transformer blocks as our baseline, leveraging its proven effectiveness for range image processing. The network employs row-based patching (\(1 \times 4\)) to preserve vertical resolution during tokenization and circular padding to handle the \(360^\circ\) continuity. Our contributions enhance this baseline through frequency-aware attention and adaptive fusion mechanisms, which are detailed in the following sections.  

The training objective minimizes the L1 loss between predicted and ground truth range images:
\begin{equation}
    L = \lVert I_h - \hat{I}_h \rVert_{1}
\end{equation}

where \(\hat{I}_h\) denotes the network prediction. The final 3D point cloud is obtained through inverse spherical projection of the predicted range image.

\subsection{Architecture Overview}
Our architecture builds upon TULIP's successful U-Net design with Swin Transformer blocks, enhancing it with two key innovations: Frequency-Aware Window Attention and Adaptive Multi-Scale Fusion, as illustrated in Figure~\ref{fig:FLASH_FA}. The encoder-decoder structure processes range images through a hierarchical feature extraction and reconstruction pipeline. 

The encoder consists of four stages, each containing paired Swin Transformer blocks followed by patch merging for \(2\times\) downsampling. Starting from the tokenized input with embedding dimension \(C=96\), each stage doubles the channel dimension while halving the spatial resolution. The decoder mirrors this structure with patch expanding layers for upsampling, progressively reconstructing the high-resolution representation. Skip connections link corresponding encoder-decoder levels, but unlike TULIP's simple concatenation, we employ our adaptive fusion module to intelligently combine multi-scale features.  

The network maintains TULIP's effective design choices: row-based patching (\(1\times4\)) preserves vertical resolution, which is critical for upsampling; circular padding handles the \(360^\circ\) horizontal continuity, and non-square windows (\(2\times8\)) better capture range image characteristics. Our approach integrates seamlessly into this architecture; the Frequency-Aware Attention substitutes the typical window attention in every Transformer block, and the Multi-Scale Fusion modules improve each skip connection.

\subsection{Frequency-Aware Window Attention}
Standard window-based attention in TULIP processes features exclusively in the spatial domain within fixed local windows. While effective in capturing local geometric patterns, this approach has a limited receptive field and may miss global structures that manifest more clearly in the frequency domain. Range images display pronounced periodic patterns from regular scanning processes and contain sharp edges at object boundaries-characteristics that are well-separated in the frequency domain.

Our Frequency-Aware Window Attention (FA) addresses this limitation through a dual-branch design that processes information in both spatial and frequency domains simultaneously. The spatial branch preserves TULIP's effective local attention mechanism, while the frequency branch captures global patterns through FFT-based processing.  

Given input features \(X \in R^{B\times H\times W\times C}\), we first apply window partitioning to obtain 
\begin{equation}
    X_w \in R^{N\times M^2\times C}
\end{equation}

where \(N\) is the number of windows and \(M\) is the window size.  
The spatial branch computes standard multi-head self-attention:
\begin{equation}
\text{Attention}_{\text{spatial}} = \text{Softmax}\!\left(\frac{QK^T}{\sqrt{d}} + B\right)V
\end{equation}
where \(Q, K, V\) are query, key, and value projections, and \(B\) represents relative position bias.  

In parallel, the frequency branch processes the global context. We extract the average feature across channels and apply 2D FFT:
\begin{equation}
X_{\text{freq}} = \text{FFT2D}(\text{mean}(X, \text{dim}=C)),
\end{equation}
\begin{equation}
F_{\text{attn}} = \sigma(\text{Conv}(|X_{\text{freq}}|)),
\end{equation}
\begin{equation}
X_{\text{freq\_out}} = \text{iFFT2D}(X_{\text{freq}} \odot F_{\text{attn}}),
\end{equation}
where \(\odot\) denotes element-wise multiplication and \(\sigma\) is the sigmoid activation. The frequency attention \(F_{\text{attn}}\) is learned through a lightweight convolutional network that identifies important frequency components.  

The final output combines both branches with a learnable weight \(\alpha\):
\begin{equation}
\text{Output} = \text{Attention}_{\text{spatial}} + \alpha \cdot \text{expand}(X_{\text{freq\_out}}),
\end{equation}
where \(\alpha\) is initialized to 0.1 and learned during training, allowing the network to balance local and global information.  

\renewcommand{\arraystretch}{1.2}
\begin{table}[t]
\centering
\caption{Ablation study demonstrating the effectiveness of Multi-Scale Fusion (MSF) and Frequency-Aware Attention (FA) components for 4× upsampling
}
\resizebox{\linewidth}{!}{%
\begin{tabular}{cc|cccccc}
    \hline
   MSF&FA&MAE $\downarrow$ & CD $\downarrow$ & IoU $\uparrow$  & Pre $\uparrow$& Re $\uparrow$&F1 $\uparrow$\\ \hline \hline
   $\times$&$\times$&0.4392& 0.1348& 0.3678& 0.5355& 0.5356&0.5355\\ \hline
   $\checkmark$&$\times$&0.3904 & 0.1209 & 0.3721  & 0.5433& 0.5368&0.5400\\ 
   $\checkmark$&$\checkmark$&\textbf{0.3899} & \textbf{0.1161} & \textbf{0.3972}  & \textbf{0.5708}& \textbf{0.5617}&\textbf{0.5661}\\ \hline
\end{tabular}
}
\label{table:ablation}
\end{table}
\renewcommand{\arraystretch}{1.0}

\subsection{Adaptive Multi-Scale Fusion}

Skip connections in U-Net architectures preserve fine-grained details during upsampling, but standard concatenation treats all features equally, regardless of their relevance or spatial location. This uniform approach is suboptimal for range images, where the importance of information varies dramatically.  

Our Adaptive Multi-Scale Fusion module (MSF) replaces simple concatenation with an intelligent fusion strategy that learns position-specific weights for features at multiple scales. Figure~\ref{fig:MSF} illustrates the detailed architecture of our MSF module. Given encoder features \(X_e\) and decoder features \(X_d\), we first align dimensions if necessary through linear projection, then extract multi-scale representations.

We apply three parallel convolutions:
\begin{equation}
\begin{aligned}
F_{1\times1} &= \text{Conv}_{1\times1}(X_{\text{combined}}), \\
F_{3\times3} &= \text{Conv}_{3\times3}(X_{\text{combined}}), \\
F_{5\times5} &= \text{Conv}_{5\times5}(X_{\text{combined}})
\end{aligned}
\end{equation}

where \(X_{\text{combined}} = X_e + X_d\).  

Instead of simple averaging, we learn adaptive weights:
\begin{equation}
\begin{aligned}
W = \text{Softmax}(\text{Conv}_{1\times1}([F_{1\times1}, F_{3\times3}, F_{5\times5}])), \\
F_{\text{fused}} = W_1 \odot F_{1\times1} + W_2 \odot F_{3\times3} + W_3 \odot F_{5\times5}.
\end{aligned}
\end{equation}

We further refine features with CBAM (Convolutional Block Attention Module) \cite{woo2018cbam}. 
Specifically, channel attention highlights informative feature channels and spatial attention focuses on important spatial locations. 

\section{Experimental Results}

\subsection{Experimental Setup}
\textbf{Dataset and Preprocessing} We evaluate our method on the KITTI dataset \cite{kitti}, which provides high-quality LiDAR data from Velodyne HDL-64E sensors. The dataset is split into 20,000 training and 2,000 test frames, ensuring no spatial overlap between splits. Following standard practice, we downsample the 64-channel data by a factor of 4 to create 16-channel input, simulating low-resolution sensor data. Range values are log-transformed using $log(r + 1)$ to compress the dynamic range and improve training stability.

\textbf{Training Configuration} We train our network using the AdamW optimizer with $\beta_1=0.9$, $\beta_2=0.999$, and weight decay of $0.01$. Training is conducted on two NVIDIA A6000 GPUs (48 GB each) with a batch size of 8. We employ cosine annealing with warm restarts for learning rate scheduling, starting with an initial learning rate of $5\times10^{-4}$. After a 60-epoch warmup, the learning rate follows cosine decay with restarts every 600 epochs, reducing the peak by 30\% each cycle.

\begin{table}[t]
\centering
\caption{Overall Performance Comparison with existing models on KITTI Dataset}
\label{table:kitti_comparison}
\begin{tabular}{l|cccc}
\hline
Method & MAE $\downarrow$ & CD $\downarrow$ & IoU $\uparrow$ & F1 $\uparrow$ \\
\hline \hline
SwinIR \cite{liang2021swinir} & 0.5776 & 0.1874 & 0.3627 & 0.5300 \\
LiDAR-SR \cite{lidar_sr} & 0.7947 & 0.2992 & 0.2089 & 0.3433 \\
TULIP \cite{yang2024tulip} & 0.4354 & 0.1342 & 0.3819 & 0.5502 \\
\textbf{FLASH (Ours)} & \textbf{0.3899} & \textbf{0.1161} & \textbf{0.3972} & \textbf{0.5661} \\
\hline
\end{tabular}
\end{table}
\renewcommand{\arraystretch}{1.0}

\renewcommand{\arraystretch}{1.2}
\begin{table*}[t]
\centering
\caption{Distance-based Performance Comparison on KITTI Dataset}
\label{tab:distance_comparison}
\begin{tabular}{l|cccc|cccc}
\hline
\multirow{2}{*}{Method} & \multicolumn{4}{c|}{Near Range (0-30m)} & \multicolumn{4}{c}{Far Range (30-60m)} \\
\cline{2-9}
 & MAE (m) $\downarrow$ & CD $\downarrow$ & IoU $\uparrow$ & F1 $\uparrow$ & MAE (m) $\downarrow$ & CD $\downarrow$ & IoU $\uparrow$ & F1 $\uparrow$ \\
\hline \hline
SwinIR \cite{liang2021swinir} & 0.299 & 0.059 & 0.394 & 0.563 & 3.064 & 2.921 & 0.092 & 0.168 \\
LiDAR-SR \cite{lidar_sr} & 0.443 & 0.118 & 0.229 & 0.371 & 4.149 & 5.347 & 0.010 & 0.019 \\
TULIP \cite{yang2024tulip} & 0.255 & 0.049 & 0.418 & 0.587 & 2.302 & 2.196 & 0.091 & 0.167 \\
\textbf{FLASH (Ours)} & \textbf{0.239} & \textbf{0.041} & \textbf{0.434} & \textbf{0.603} & \textbf{2.045} & \textbf{1.911} & \textbf{0.105} & \textbf{0.189} \\
\hline
\end{tabular}
\end{table*}
\renewcommand{\arraystretch}{1.0}

\textbf{Evaluation Metrics} We assess performance using both 2D and 3D metrics: Mean Absolute Error (MAE) for range image quality, Chamfer Distance (CD) for 3D geometric accuracy, and Intersection over Union (IoU) for volumetric consistency after voxelization with $0.1m$ resolution. Additionally, we report Precision (Pre), Recall (Re), and F1-score to evaluate point cloud completeness and accuracy.

\textbf{Baselines} We compare against three state-of-the-art LiDAR super-resolution methods: (1) TULIP, our baseline architecture with standard window attention and simple concatenation fusion; (2) LiDAR-SR, a CNN-based U-Net; (3) SwinIR, a leading transformer-based image super-resolution method adapted for range images. All methods are trained on the same dataset splits with their recommended configurations.

\subsection{Ablation Studies}
\renewcommand{\arraystretch}{1.2}

\renewcommand{\arraystretch}{1.2}
\begin{table}[b]
\centering
\caption{Comparison with MC Dropout Enhancement}
\label{tab:mc_dropout}
\resizebox{\linewidth}{!}{%
\begin{tabular}{l|cccccc}
\hline
Method & \makecell{MC\\Dropout} & MAE $\downarrow$ & CD $\downarrow$ & IoU $\uparrow$ & F1 $\uparrow$ & Time (ms) $\downarrow$ \\
\hline \hline
TULIP & $\times$ & 0.4354 & 0.1342 & 0.3819 & 0.5502 & 14 \\
TULIP & $\checkmark$ & 0.4070 & 0.1250 & 0.3853 & 0.5538 & 134 \\
\textbf{FLASH (Ours)} & $\times$ & \textbf{0.3899} & \textbf{0.1161} & \textbf{0.3972} & \textbf{0.5661} & 15 \\
\hline
\end{tabular}
}
\begin{flushleft}
{\footnotesize *MC Dropout: 20 samples processed in batches of 8.}
\end{flushleft}
\end{table}
\renewcommand{\arraystretch}{1.0}

We conduct comprehensive ablation studies to validate the effectiveness of each component in our architecture. Table \ref{table:ablation} presents the quantitative results on the KITTI test set.

\textbf{Multi-Scale Fusion (MSF) }Replacing TULIP's simple concatenation with our MSF module yields substantial improvements. The MAE decreases to 0.3904 while IoU reaches 0.3721, demonstrating better volumetric reconstruction. The adaptive weighting mechanism effectively combines features at different scales, particularly benefiting object boundaries where appropriate scale selection is crucial.

\textbf{Frequency-Aware Attention (FA)} Adding FA to MSF further enhances performance across all metrics. The IoU improves from 0.3721 to 0.3972, indicating superior geometric accuracy. More notably, precision increases to 0.5708 while maintaining recall at 0.5617, suggesting that frequency domain processing helps eliminate spurious predictions without losing valid points.

The results demonstrate clear synergy between components: MSF provides adaptive local feature combination while FA adds global context through frequency analysis. This complementary relationship is evident in the improved Chamfer Distance (0.1209 to 0.1161) and balanced F1-score (0.5661), confirming more accurate and complete point cloud reconstruction. The consistent gains across both 2D and 3D metrics validate that our dual-domain approach addresses fundamental limitations in existing methods.

\begin{figure*}[h]
    \centering
    \includegraphics[width=\textwidth]{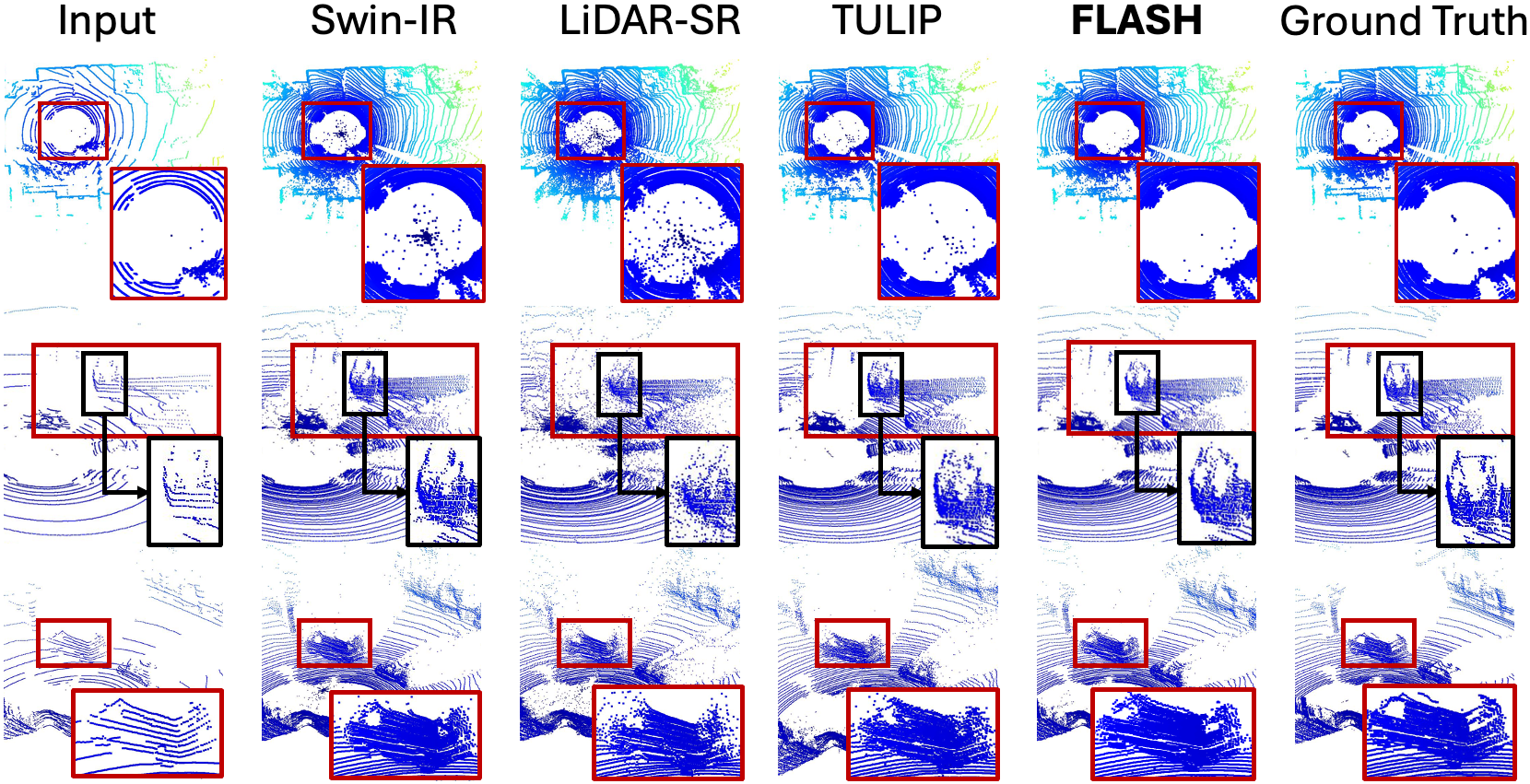}
    \caption{Qualitative results on KITTI. Comparison of super-resolution methods showing (top) noise suppression near sensor, (middle) edge preservation on large vehicles, and (bottom) Fine structural detail recovery on a van's rear section. FLASH demonstrates superior performance in preserving geometric sharpness and reducing artifacts compared to SwinIR, LiDAR-SR, and TULIP.}
    \label{fig:qualitative}
\end{figure*}

\subsection{Comparison with Existing Models}
We evaluate FLASH against existing LiDAR super-resolution methods on the KITTI test set, analyzing both overall performance and distance-specific behavior to understand model characteristics across different range intervals.

\subsubsection{Overall Performance}
Table~\ref{table:kitti_comparison} presents quantitative results comparing FLASH with three representative methods: TULIP (our baseline), SwinIR (adapted from image super-resolution), and LiDAR-SR (CNN-based approach).

FLASH achieves the best performance across all metrics. Compared to LiDAR-SR, it reduces MAE by 51\% and nearly doubles IoU, highlighting the limitations of CNN architectures in preserving sharp discontinuities. Against TULIP, FLASH shows consistent improvements (10\% lower MAE and 13\% lower CD), demonstrating the effectiveness of our architectural enhancements over the baseline transformer design.

The performance gains stem from the complementary nature of our components. While both FLASH and TULIP share a transformer backbone, our dual-domain processing enables better capture of both local details and global patterns. The frequency-aware attention expands the effective receptive field beyond window boundaries, while adaptive fusion preserves geometric details through learned multi-scale aggregation. This synergy is evident in the balanced improvements across error metrics (MAE), geometric fidelity (CD), and volumetric accuracy (IoU, F1).

All methods are evaluated using single-pass deterministic inference to assess core architectural performance. These results confirm that frequency domain processing and adaptive fusion effectively address the core challenges of LiDAR super-resolution.

\subsubsection{Distance-based Analysis}
To better understand model behavior across varying ranges, we evaluate performance in two distinct intervals: near range (0-30m) and far range (30-60m). This analysis reveals how different architectures handle the increasing sparsity and noise characteristics at longer distances.

Table~\ref{tab:distance_comparison} presents the distance-based evaluation results. All methods exhibit significant performance degradation at far ranges, with MAE increasing by approximately 8-10 times and IoU dropping to below 0.11 for all approaches. This significant drop illustrates the primary difficulty with LiDAR super-resolution: as the distance increases, point density diminishes dramatically, and the measurement noise increases.

In the near range, FLASH achieves the lowest MAE of 0.239 meters, representing a 6.3\% improvement over TULIP and reducing the error of LiDAR-SR by nearly half. This improvement suggests that frequency-aware attention effectively captures the periodic scanning patterns prevalent in dense regions. Our method maintains an IoU of 0.434, demonstrating enhanced geometric consistency in reconstructing nearby objects.

The far range presents greater challenges, with all methods showing significant performance degradation. FLASH demonstrates the best resilience, achieving an MAE of 2.045 meters (an 11\% improvement over TULIP and less than half of the error of LiDAR-SR) Notably, the Chamfer Distance values increase dramatically at far ranges, particularly for LiDAR-SR (5.347), indicating substantial geometric distortion in distant regions.

The F1 scores reveal comparable degradation patterns between FLASH and TULIP: both experience approximately 3 times reduction from near to far range. However, FLASH maintains a 13\% higher F1 score at far range, suggesting that the frequency-aware attention and multi-scale fusion provide meaningful benefits in challenging conditions. This consistent advantage across all distance-based metrics demonstrates the effectiveness of our architectural enhancements over the baseline.

\subsubsection{Comparison with Uncertainty-Enhanced Baselines}
Table~\ref{tab:mc_dropout} presents a critical comparison between FLASH and TULIP with Monte Carlo Dropout enhancement. While MC Dropout improves TULIP's performance, it requires 20 forward passes, increasing inference time from 14 ms to 134 ms. Despite this 10× computational overhead, TULIP with MC Dropout still underperforms FLASH's single-pass inference, which achieves a lower MAE and a higher IoU and F1 score in only 15 ms.

Our model demonstrates robust predictions without explicit uncertainty modeling. The frequency domain suppresses noise, and adaptive fusion weights features based on context, providing implicit uncertainty handling. FLASH reaches 66 FPS versus 7.5 FPS for TULIP with MC Dropout, showing that well-designed architecture can surpass stochastic methods in both accuracy and efficiency.

\subsection{Qualitative Analysis}

Figure~\ref{fig:qualitative} presents visual comparisons across three challenging scenarios from the KITTI test set.

The first row illustrates noise suppression around the sensor mounting region. While LiDAR-SR generates numerous spurious points and TULIP still exhibits scattered artifacts, FLASH produces a clean reconstruction through adaptive multi-scale fusion. This demonstrates FLASH's inherent noise suppression without requiring Monte Carlo Dropout, achieving superior quality in a single deterministic pass.

The second row highlights edge preservation on large vehicles. SwinIR and LiDAR-SR show substantial blurring along vehicle boundaries, while TULIP smooths vertical surfaces. In contrast, FLASH preserves sharp discontinuities through frequency-aware attention, maintaining structural integrity that directly contributes to the improved IoU metrics.

The third row evaluates fine detail recovery in a van's rear window area. LiDAR-SR produces amorphous regions, and TULIP loses window frame distinction through over-smoothing. FLASH successfully delineates window frames by leveraging multi-scale fusion; smaller receptive fields capture thin boundaries, while larger ones provide contextual consistency.

These qualitative results confirm our quantitative findings: FLASH consistently reduces noise, preserves boundaries, and recovers fine structures more faithfully than competing methods, directly improving object detection accuracy for autonomous driving applications.

\section{Conclusion}
We presented FLASH, a novel LiDAR super-resolution framework that successfully adapts frequency domain processing to 3D perception tasks. Our dual-domain approach, combining Frequency-Aware Window Attention with Adaptive Multi-Scale Fusion, addresses the fundamental limitations of existing methods.

Our experiments reveal key insights. First, the synergy between MSF and FA is critical: MSF handles local geometric details through adaptive scale selection, while FA captures global patterns via frequency analysis, together achieving superior reconstruction quality. Second, FLASH surpasses TULIP with Monte Carlo Dropout in accuracy while maintaining single-pass inference, proving that architectural design can replace computationally intensive stochastic methods. 

The consistent performance across all distance ranges validates our dual-domain approach. By successfully transferring frequency domain techniques from 2D vision to LiDAR processing, this work opens new directions for practical 3D perception in resource-constrained autonomous systems.

\addtolength{\textheight}{-12cm}   




%

\bibliographystyle{IEEEtran}
\bibliography{IEEEabrv,mybibfile}

\end{document}